\documentclass{article}

\pdfoutput=1

\usepackage[preprint]{neurips_2024}

\usepackage[utf8]{inputenc} 
\usepackage[T1]{fontenc}    
\usepackage{hyperref}       
\usepackage{url}            
\usepackage{booktabs}       
\usepackage{amsfonts}       
\usepackage{nicefrac}       
\usepackage{microtype}      
\usepackage{xcolor}         
\usepackage{multirow} 
\usepackage{mathrsfs}
\usepackage{times}
\usepackage{soul}
\usepackage{url}
\usepackage{graphicx} 
\usepackage[ruled,vlined]{algorithm2e}
\usepackage[utf8]{inputenc}
\usepackage{caption}
\usepackage{amsmath}
\usepackage{amsthm}
\usepackage{booktabs}
\usepackage{verbatim}

\usepackage{array} 

\title{
Redundant feature screening method for human activity recognition based on attention purification mechanism}

\author{Xiaoyang Li , Yixuan Jiang , Junze Zhu , Haotian Tang , Dongchen Wu , Hanyu Liu , Chao Li
}
\begin{document}

\maketitle

\begin{abstract}
In the field of sensor-based Human Activity Recognition (HAR), deep neural networks provide advanced technical support. Many studies have proven that recognition accuracy can be improved by increasing the depth or width of the network. However, for wearable devices, the balance between network performance and resource consumption is crucial. With minimum resource consumption as the basic principle, we propose a universal attention feature purification mechanism, called MSAP, which is suitable for multi-scale networks. The mechanism effectively solves the feature redundancy caused by the superposition of multi-scale features by means of inter-scale attention screening and connection method. In addition, we have designed a network correction module that integrates seamlessly between layers of individual network modules to mitigate inherent problems in deep networks. We also built an embedded deployment system that is in line with the current level of wearable technology to test the practical feasibility of the HAR model, and further prove the efficiency of the method. Extensive experiments on four public datasets show that the proposed method model effectively reduces redundant features in filtered data and provides excellent performance with little resource consumption.

\end{abstract}

\section{Introduction}
We are witnessing substantial growth in affordable, reliable, and proficient solutions as we move towards the era of autonomous systems. One such application is Human Activity Recognition (HAR), a domain which utilizes AI technology to unearth motion behavior patterns \citet{Asurvey}. HAR, with its ability to classify human activity signals into different actions, has profound economic and research implications, particularly in intelligent healthcare. Predominantly, there exist two categories of HAR systems: vision-based and sensor-based \citet{AReview}. Whereas vision-based methods involve image processing and are susceptible to various environmental factors, sensor-based techniques are more reliable and cost-effective, and less vulnerable to external disturbances. Sensor-based HAR systems find extensive usage in various applications such as health monitoring, fall detection, athlete tracking, and electrocardiogram analysis \citet{TII_ECG}. 

Classical machine learning methods such as decision trees (DT), support vector machines (SVM), random forests (RF), and naive Bayes (NB) have found widespread usage in initial sensor-based HAR tasks due to their low computational complexity and smaller dataset suitability \citet{Z.Wang_D.Wu,Z.Chen}. However, the inability of these conventional methods to capture complex relationships has necessitated the usage of deep learning methods. Progress in mobile sensing technology has eased the deployment of deep neural networks such as Convolutional Neural Networks, Recurrent Neural Networks, and Transformers in HAR \citet{M.Zeng_and_Nguyen,B.Meng,X.Li_Y.Wang}. These deep learning models, endowed with robust feature learning and complex temporal relationship modeling, thus offer distinct improvements in sensor-based HAR tasks over traditional methods.

Nonetheless, challenges exist in sensor-based HAR: 

\textbf{Difficulty balancing performance and efficiency:}
HAR is essentially a classification task. Despite successful feature extraction endeavours, issues persist in practical application; for example, confusion when learning similar action features, resulting in particular action categories being indistinguishable, or difficulty in applying individual learned features to others \citet{Overview_Challenges_Opportunities}, thereby making it a challenging task. Common strategies to mitigate these issues include incorporating LSTM, Transformer, and other modules adept at extracting sequence features based on the convolutional network \citet{TransSleep,F.J.Ordonez}; escalating the depth, width, and parameter quantity of the network \citet{S.Mekruksavanich}; utilizing traditional Time series network models for feature extraction \citet{Y.Guan,M.Zeng}. These methods frequently consume copious resources, posing challenges for wearable devices to maintain high efficiency with high accuracy. Some people \citet{NNU2023} have proposed using embedded deployment experiments to simulate actual use conditions, but the performance of the equipment used in most current deployment methods in HAR is generally weaker than that of actual application equipment, and the experimental design is too simple. Experiments and discussions on efficiency, complexity, and inference latency are not comprehensive.

\textbf{Noise Interference:}
The accuracy of recognition in sensor-based HAR heavily relies on features extracted from raw signals which are often polluted by noise \citet{Overview_Challenges_Opportunities}, thereby magnifying the challenge of feature extraction. During data acquisition, sensors can produce random noise due to technical errors and accuracy constraints. Furthermore, the preprocessing stage might incorporate irrelevant information not linked to the actual signal, generating noise. Environmental noise and electromagnetic interference can also taint data. Traditional noise reduction methods such as Gaussian filters and wavelet filters are extensively used to process signal data amassed by sensors like ECGs. However, processing different movements through noise reduction methods, such as three-axis acceleration information, becomes challenging \citet{S.-M.Lee}. Currently, there is no unified method for denoising signals from multiple modes of homologous heterogeneous sensors. Most filtering methods are only suitable for processing noise in a specific frequency range, resulting in the inevitable loss of homologous related information when applying traditional methods.

This paper, therefore, makes three main contributions:

\begin{enumerate}
 \item We introduce a lightweight attention purification module(MSAP) for multi-scale networks that can easily integrate with any multi-scale network, offering scalability and versatility. It efficiently curtails the network's overprocessing of redundant features with minimal resource consumption.
 \item We enhance the soft threshold selection method of the residual shrinkage network to screen redundant features more proficiently and propose DRSN-M, which is optimal for HAR systems.
 \item We built a comprehensive and detailed wearable deployment simulation system using Raspberry Pi 5 devices that meet today's wearable functions. On the basis of completing basic experiments, we further analyzed the model's complexity and inference latency during actual deployment. 
\end{enumerate}

\section{Related work} \label{Related work}
\subsection{Feature extraction and computational efficiency}
The advent of myriad deep learning architectures in recent years has incited the formation of an array of resilient deep learning methods for HAR. Particularly, models employing Convolutional Neural Networks (CNN) and Long Short-Term Memory (LSTM) have demonstrated potent results \citet{L.MinhDang,M.Zeng_and_Nguyen}. Presently, the pivotal Multi-headed Self-attention mechanism within the Transformer structure has earned several leading-edge performances in this field. However, these methods also result in greater computational overhead.\textbf{ }Gao et al. \citet{DanHAR} innovated a unique dual attention mechanism in their DanHAR approach, thereby presenting a framework that consolidates CNN channel attention and temporal attention. This intensifies the attention toward essential sensor patterns and time steps, culminating in substantial enhancements to the interpretability of Multimodal HAR. Cheng et al. \citet{NNU2023} conduct actual deployment experiments through Raspberry Pi devices to prove the practical feasibility of their method. This is where the idea for our deployment system came from.

\subsection{Denoising Methods for HAR}
Numerous denoising methods are widely adopted in the domain of signal sensors, including low-pass filters, mean filters, linear filters, and Kalman filters. These methods display satisfactory performance for activity signals \citet{Overview_Challenges_Opportunities}. However, finding a universal filtering method that fits all circumstances is challenging in HAR systems, as the noise patterns for varied activities and sensors diverge significantly. Human activity encompasses a vast frequency range, extending from low-frequency everyday activities to high-frequency hand movements, all of which carry vital information. Zhang et al. \citet{Y.Zhang_EMG} formulated the Noise-Assisted Multivariate Empirical Mode Decomposition (NA-MEMD) method, intended for preprocessing multi-channel electromyogram (EMG) signals to extract statistical data delineating the temporal and spatial attributes of diverse muscle groups. Vijayvargiya \citet{AnkitVijayvargiya} put forth a wavelet denoising approach using the Wavelet Decomposition-Enhanced Empirical Mode Decomposition (WD-EEMD) preprocessing technology to eliminate noise from the SEMG signals of calf muscles for activity detection. 

\begin{figure}
 \centering
 \includegraphics[width=0.8\linewidth]{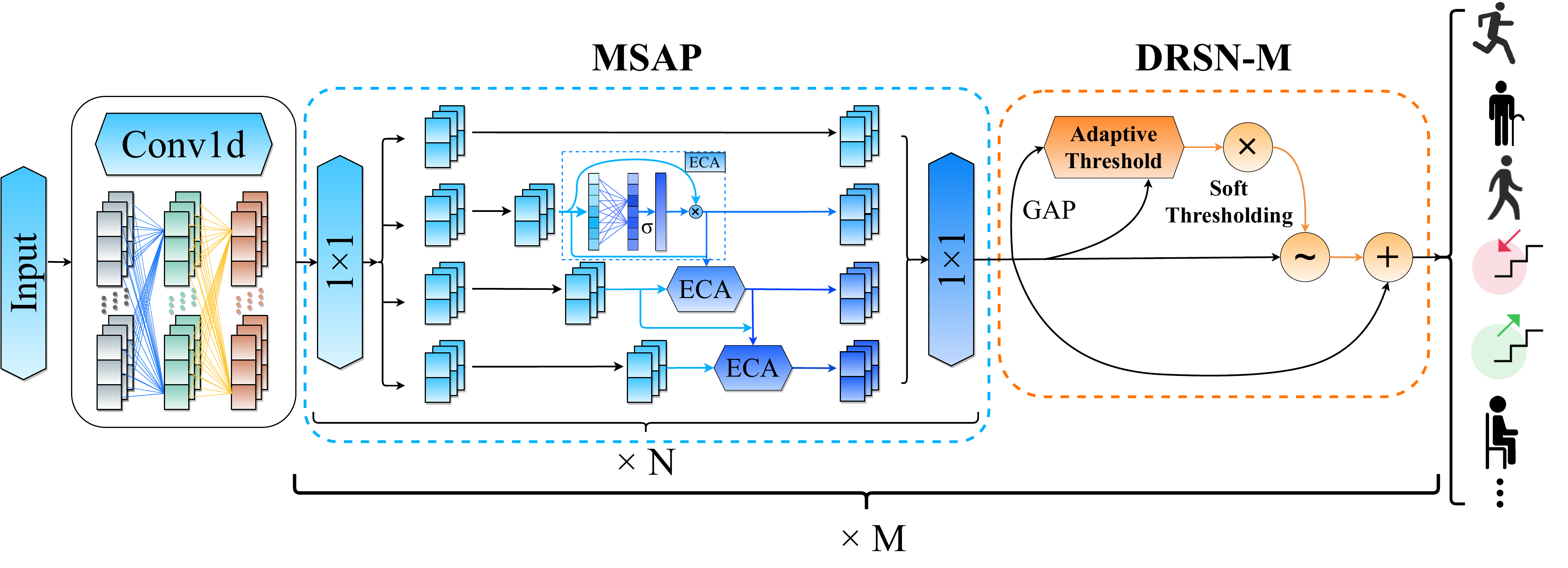}
 \caption{The method proposed in this paper is the total process of human activity identification}
 \label{fig1:ALL}
\end{figure}
\section{Methods} \label{methods}
The purpose of this paper is to demonstrate the excellent performance of the attention purification mechanism and the interlayer noise reduction network architecture by constructing a multi-scale CNN network. Our aim is to achieve higher performance in the sensor HAR system while consuming fewer resources.  In a standard HAR task, we first need to process the raw signal data, which may be inertial sensor signals such as three-axis acceleration and angular velocity, or ECG or EMG signals.We utilize the sliding window method to segment the signal into overlapping windows. While we do not perform any noise reduction in the pre-processing stage, we incorporate this step in the feature extraction stage. 

Specify the input feature $x_\tau\in R^{C\times T}$, where $C$ is the feature channel and $T$ is the channel time length. Input feature $x_\tau$ first undergoes preliminary feature extraction through a set of regular one-dimensional convolutions. In order to capture effective features at multiple scales, we propose an MSAP structure that can be used for multi-scale networks in the feature extraction stage, which can easily adapt to various multi-scale networks and provide a more efficient connection method for complex networks. Split the feature map into $s$ subsets of feature maps with the same channel size $x_i$, for the first subset of feature maps, we use convolution $K_i()$ and attention $A_i()$ for feature optimization, and then transfer the features to the next scale according to Figure 1. For $x_i$, when passing in the next scale, the $x_i$ and $x_{i+1}$ are superimposed and the features are filtered in the ECA attention, which can eliminate unnecessary redundant features of the scale. At the same time, $x_i$ is output in its own scale, which we define as $y_i$, The final ideal scale expression is $y_i = A_i[K_i(x_i)+K_{i-1}(x_{i-1})+x_{i-1}]$. All outputs are then concatenated and passed into a $1\times1$ convolution. The DRSN module is added between MSAP modules with different channel sizes, which we define as DRSN-Modular(DM). In DM , we first decompose the input feature, which is mainly done by convolution, then filter all the decomposed features within the threshold, and finally reconstruct all the filtered signals. Among them, the threshold is set one by one by the ECA for each channel of the feature. During the construction process, we refer to a number of networks about multiscale, attention, and residual structures. The result is a structure that is more efficient and capable of extracting features, while maintaining a complex MSAP and MSAP-DM network structure similar to the original structure.

\subsection{Multi-scale Attention Feature Extraction}
We designed an attention purification mechanism based on a multi-scale residual network. Firstly, we use a simple multi-scale residual network as the basic framework. Then, we added inter-channel correlations between each scale to enable the model to better focus on the combined information of features at multiple scales. Finally, we incorporated an attention purification mechanism to reduce redundant features across multiple scales. Our model effectively captures important features at various scales while avoiding unnecessary and difficult-to-process features that exist at multiple scales. As shown in Figure \ref{fig1:ALL}. After the $1 \times 1$ convolution, we evenly split the feature maps into $s$ feature map subsets, denoted by $x_i$, where $i \in \{1, 2, \dots, s\}$. Each feature subset $x_i$ has a channel size equal to $1/s$ of the input feature map. Except for $x_1$, each $x_i$ has a corresponding $3 \times 1$ convolution, denoted by $K_i()$. We denote the output of $K_i()$ as $y_i$. The feature subset $x_i$ is added to the output of $K_{i-1}()$ and then fed into $K_i()$. To reduce parameters while increasing $s$, we omit the $3 \times 1$ convolution for $x_1$. We then capture features of different scales and process them using channel attention, denoted as $A_i()$. At the next scale, we combine the pre-processed features $K_{i-1}(x_{i-1})$ and the post-processed features $y_{i-1}$ with the post-processed features $K_i(x_i)$. This process continues until all scales of features have been processed. The formula for $y_i$ in this model is:
\begin{equation}
\begin{aligned}
y_i=\begin{cases}
x_i, & i=1\\ 
A_i[K_i(x_i)], & i=2\\
A_i[K_i(x_i)+K_{i-1}(x_{i-1})+y_{i-1}], & 2<i\leq s.
\end{cases}
\end{aligned}
\end{equation}
This way, it preserves the original information of the features, and to some extent strengthens the weight of important features. When using new attention, it can also further filter the current situation, reducing the relatively ineffective parts in the features that were previously given higher weight, and improving the relatively effective parts in the features that were previously given lower weight. This approach can not only select effective features more accurately but also alleviate the problem of model overfitting.

\subsection{Noise reduction module}
\begin{figure}[htbp]
  \centering
  \begin{minipage}[b]{0.53\textwidth}
    \includegraphics[width=\textwidth]{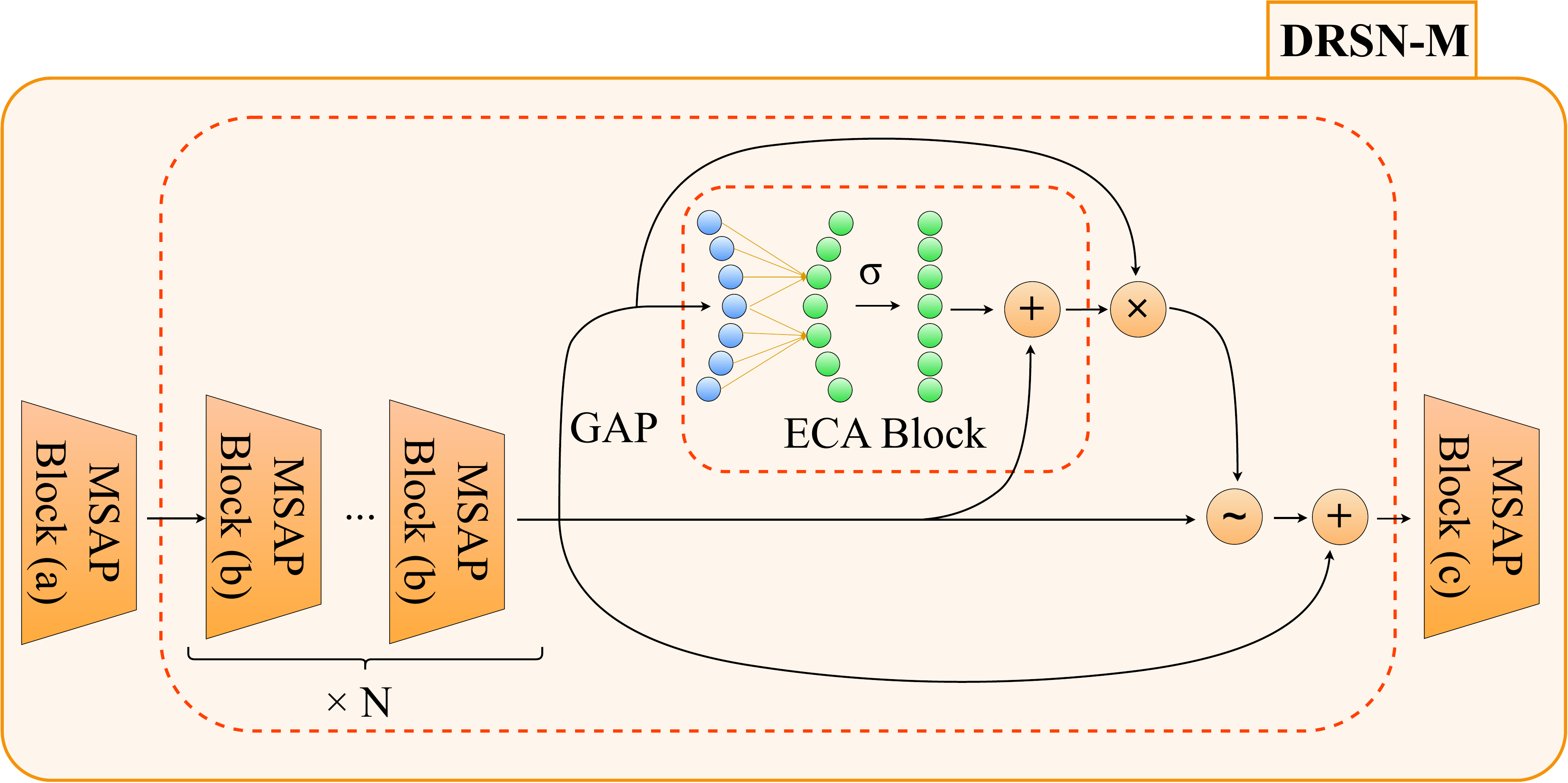}
    \caption{The soft threshold module is used only between MSAP groups}
    \label{fig3:DM}
  \end{minipage}
  \hfill 
  \begin{minipage}[b]{0.40\textwidth}
    \includegraphics[width=\textwidth]{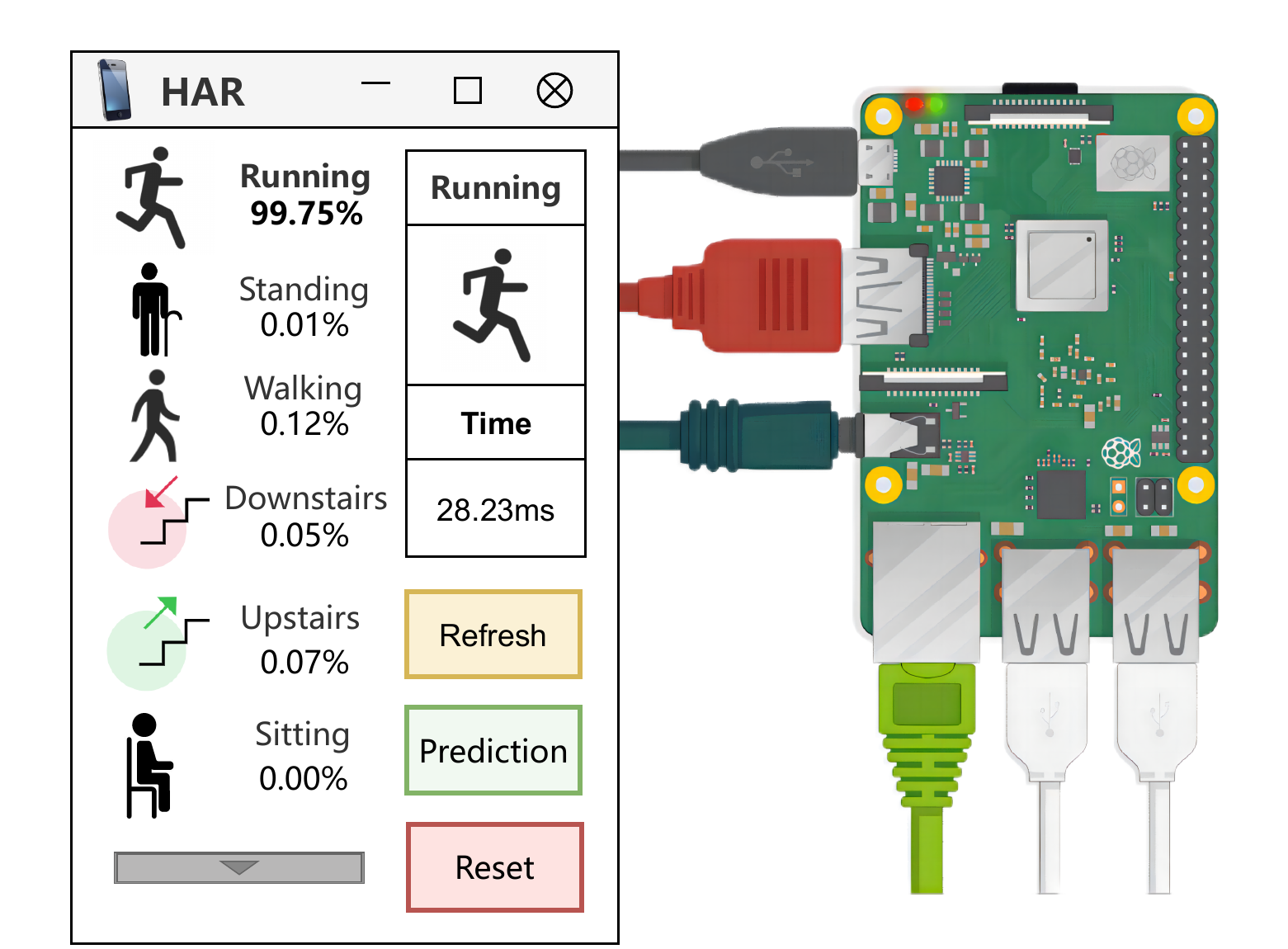}
    \caption{Model deployment}
    \label{fig:Model deployment}
  \end{minipage}
\end{figure}
Based on the Deep Residual Shrinkage Network (DRSN), we proposed a denoising network DRSN-M that can handle redundant features more efficiently, and successfully introduced it into the field of sensor-based HAR to solve the noise problem. The soft thresholding method has always been a key step in signal denoising \citet{AnkitVijayvargiya}. Generally, the original signal is transformed into a domain where near-zero digital features are not important, and then the soft threshold is applied to convert near-zero features to zero. The soft threshold function can be represented as:
\begin{equation}
\begin{aligned}
y_i=\begin{cases}
x-\tau, & x>\tau\\ 
0, & -\tau \leq x \leq \tau\\
x + \tau,& x<-\tau. 
\end{cases}
\end{aligned}
\end{equation}
where \(x\) is the input feature, \(y\) is the output feature, and \(\tau\) is the threshold, which is a positive parameter. 

Zhao et al. \citet{M.Zhao_DRSN} proposed two structural modules: DRSN-CS and DRSN-CW. In general, DRSN-CW performs better because it allows attention to set thresholds for each channel individually. However, in deep residual networks, it is difficult to avoid the loss of effective features with a large number of denoising modules. Therefore, we have built a DRSN-M (DM) module, which is also based on the residual network structure. We define the MSAP network layer with the same convolutional kernel as an MSAP group. Each MSAP group uses a DM module for denoising. In order to reduce the impact of data dimensionality reduction on network resource consumption, we use a lightweight non-redundant ECA attention mechanism instead of the traditional SE attention mechanism and set parameters for feature processing according to the corresponding number of channels \citet{Q.Wang_ECA}. In summary, our model can solve the problem of denoising difficulties in HAR and the problem of the direct application of traditional DRSN networks. We combine it with the MSAP to form the MSAP-DM (MSAP \& DRSN-M).

\subsection{Efficient Channel Attention}
We have previously mentioned the ECA module several times, which is an efficient and lightweight attention mechanism. Compared to traditional attention mechanisms, ECA does not require complex dimensionality reduction and expansion processes\citet{{wang2020eca}}. 

The ECA module adaptively calculates the kernel size \( k \) for a 1D convolution based on the number of channels \( C \). The formula for calculating the kernel size is as follows:
\[
k = \left|\frac{\log_2(C)}{\gamma} + \frac{b}{\gamma}\right|_\mathrm{odd}
\]
This formula is used to determine the kernel size \( k \) for the 1D convolution, where \( C \) is the number of input feature channels, and \( \gamma \) and \( b \) are hyperparameters. Taking the absolute value and rounding down to the nearest $\mathrm{odd}$ number ensures that the kernel size is odd. Once the kernel size \( k \) is obtained, the ECA module applies a 1D convolution to the input features to learn the importance of each channel relative to the others. 

\subsection{Embedded device deployment}
In recent years, many studies only use high-performance device testing methods, and when actually deploying HAR models, it is often necessary to consider various limitations of the deployed devices. In view of this, we built an embedded deployment system that is in line with the current level of wearable technology to test the practical feasibility of various methods. This idea was inspired by Zhang et al.. Their deployment experiments were based on Raspberry Pi 3 and 4. Compare the inference delay time of the proposed module and the baseline model \textbf{\citet{DanHAR,CE} }. We hope to build a more comprehensive and detailed set of deployment experiments, We put the specific implementation details in the Actual deployment in the experimental part.

\section{Experiment} \label{Experiment}

This section offers a comprehensive elucidation of the conducted experiments and specificities associated with them. In order to ensure an objective evaluation of our methodology, several pertinent aspects of these employed datasets have been outlined below. 

The PAMAP2 dataset\citet{A.ReissandD.Stricker}, publicly available on the UCI repository, captures 18 diverse physical activities from nine subjects. These subjects wore three Inertial Measurement Units (IMUs) on their dominant wrist, chest, and ankle. The WISDM dataset\citet{J.R.Kwapisz}, a notable HAR benchmark from the Wireless Sensor Data Mining Lab, includes six data attributes: user, activity, timestamp, and x, y, z accelerations. The data were collected from 29 volunteers performing activities such as walking, jogging, and stair climbing using an Android smartphone. The OPPORTUNITY dataset\citet{R.Chavarriaga} meticulously documents the activities of 12 subjects in a sensor-enriched environment. The dataset, simulating a real-life setting, includes data from 15 networked sensor systems with a total of 72 sensors of 10 different types. The UCI-HAR dataset\citet{A.Ignatov} consists of sensor recordings from 30 subjects performing routine activities. The data were collected using a waist-mounted smartphone, capturing three-axis linear acceleration and three-axis angular velocity signals at a constant rate of 50 Hz.Before the datasets could be utilized for training, validation, and testing, they underwent rigorous preprocessing. 
\subsection{Evaluation Metrics}
To evaluate the performance of the proposed model for HAR, the followed metrics were used for evaluation generally.

\begin{equation}
\begin{aligned}
\text{Accuracy} &= \frac{\mathrm{TP} +\mathrm{TN}}{\mathrm{TP}+\mathrm{FN}+\mathrm{FP}+\mathrm{TN}}\\
\text{F1-macro} &= \frac{2 \times (\text{Precision} \times \text{Recall})}{\text{Precision} + \text{Recall}}\\
\text{F1-weighted} &= \sum_{i} \frac{2 \times \omega_i \times (\text{Precision}_i \times \text{Recall}_i)}{\text{Precision}_i + \text{Recall}_i}
\end{aligned}
\end{equation}

where $\mathrm{TP}$ and $\mathrm{TN}$ are the number of true and false positives, respectively, and $\mathrm{FN}$ and $\mathrm{FP}$ are the number of false negatives and false positives. $\omega_i$ is the proportion of samples of class $i$.
\subsection{Experimental environment and hyperparameters}
Tables \ref{tab:Experimental environment } and \ref{tab1:Dataset Processing Details} show the experimental environment and details of our processing of data sets and model parameters.

\begin{minipage}{\textwidth}
\begin{minipage}[c]{0.4\textwidth}
\makeatletter\def\@captype{table}
\caption{Experimental Environment }
\label{tab:Experimental environment }

\begin{tabular}{c|>{\centering\arraybackslash}p{0.27\linewidth}>{\centering\arraybackslash}p{0.27\linewidth}}
\toprule
\
\textbf{Work}& Base& Deploy\\
\hline
\textbf{PyTorch} & 2.0.0 & 2.2.2\\
\hline
\textbf{Python} & 3.8& 3.11.2\\
\hline
\textbf{Cuda} & 11.8 & N/A\\
\hline
\textbf{GPU} & RTX 4090 24G& N/A\\
\hline
\textbf{CPU} & Xeon(R) Gold 6430& Quad-core Cortex-A76\\
\hline
Memory & 120GB & 8GB\\
\bottomrule

\end{tabular}

\end{minipage}
\begin{minipage}[c]{0.65\textwidth}
\makeatletter\def\@captype{table}
\caption{Processing Details}
\label{tab1:Dataset Processing Details}
\begin{tabular}{c|>{\centering\arraybackslash}p{0.15\linewidth}>{\centering\arraybackslash}p{0.15\linewidth}>{\centering\arraybackslash}p{0.15\linewidth}c}
\toprule
\textbf{Datasets} & PAMA.& WISDM & OPPO.& UCI.\\
\hline
\textbf{Sensor} & 40 & 3 & 72 & 9 \\
\hline
\textbf{Rate} & 33Hz & 20Hz & 30Hz & 50Hz \\
\hline
\textbf{Subject} & 9 & 29 & 12 & 30 \\
\hline
\textbf{Class} & 12 & 6 & 18 & 6 \\
\hline
\textbf{Window}& 171 & 90 & 113& 128 \\
\hline
\textbf{Tr/Va/Te} & 8:1:1 & 7:2:1 & 7:2:1 & 8:1:1 \\
\hline
 \textbf{Batch size} & 128& 512 & 256 &512\\
 \hline
 \textbf{Width} & 10& 8 & 4&8\\
 \hline
 \textbf{Scale} & 4 & 4 & 12&8\\
 \hline
 \textbf{Lr} & 0.001& 0.001 & 0.001 &0.001 \\
 \bottomrule
\end{tabular}

\end{minipage}
\end{minipage}

\subsection{Ablation Experiment}
\begin{table}
\centering
\caption{Ablation Experiment}
\label{tab4:Ablation Experiment}
\begin{tabular}{cccccccc} 
\toprule
\multicolumn{2}{c}{\textbf{Model}} & \multicolumn{3}{c}{\textbf{WISDM}} & 
\multicolumn{3}{c}{\textbf{OPPORTUNITY}} \\
\cmidrule(r){1-2}\cmidrule(r){3-5} \cmidrule(r){6-8}
\textbf{Network} & \textbf{Extra Addition} & \textbf{Accuracy}& \textbf{F1-m}& \textbf{F1-w}& \textbf{Accuracy}& \textbf{F1-m}& \textbf{F1-w}\\
\cmidrule(r){1-2}\cmidrule(r){3-5} \cmidrule(r){6-8}
\multirow{3}{*}{Base}& \textbackslash{} & 96.03 & 94.47 & 96.01 & 82.43 & 77.08 & 83.69 \\
 & Scale Connections & 95.51 & 93.69 & 95.57 & 82.40 & 77.57 & 84.63 \\
 & Attention Purification& \textbf{96.85}& \textbf{95.44}& \textbf{96.89}& \textbf{84.39}& \textbf{78.35}& \textbf{84.90} \\
\cmidrule(r){1-2}\cmidrule(r){3-5} \cmidrule(r){6-8}
\multirow{2}{*}{MSAP} & DRSN & 96.46 & 95.02 & 96.41 & 81.56 & 75.02 & 82.57 \\
 & \textbf{DRSN-M(DM)}& \textbf{98.24} & \textbf{97.09} & \textbf{98.14}& \textbf{87.55} & \textbf{82.17} & \textbf{87.60} \\
\bottomrule
\end{tabular}\end{table}

We study the performance of the corresponding model and the feasibility of method combinations by setting different module combinations. Models using simple uncorrelated multi-scale networks have surpassed general-purpose neural networks like ResNet in accuracy. However, once scale interconnections are added, the accuracy decreases. We believe this is due to unnecessary feature stacking at multiple scales. In our attention-purified network, the performance is significantly improved over the former, which should be attributed to the attention-based feature selection mechanism. We believe this is due to feature redundancy caused by excessive feature stacking at multiple scales, which requires an attention purification mechanism to eliminate these features. It can be clearly seen that the performanceis significantly improved compared to the former. In addition, directly applying the DRSN noise reduction module to the MSAP does not produce good results, and the network performance is lower than the original MSAP. The DRSN-M(DM) we built solves this problem well by adjusting the noise reduction structure. On the WISDM dataset, the accuracy improves by about 1.5\% and on the OPPORTUNITY dataset by about 3\%.
\begin{figure}[htbp]
  \centering
  \begin{minipage}[b]{0.46\textwidth}
    \includegraphics[width=\textwidth]{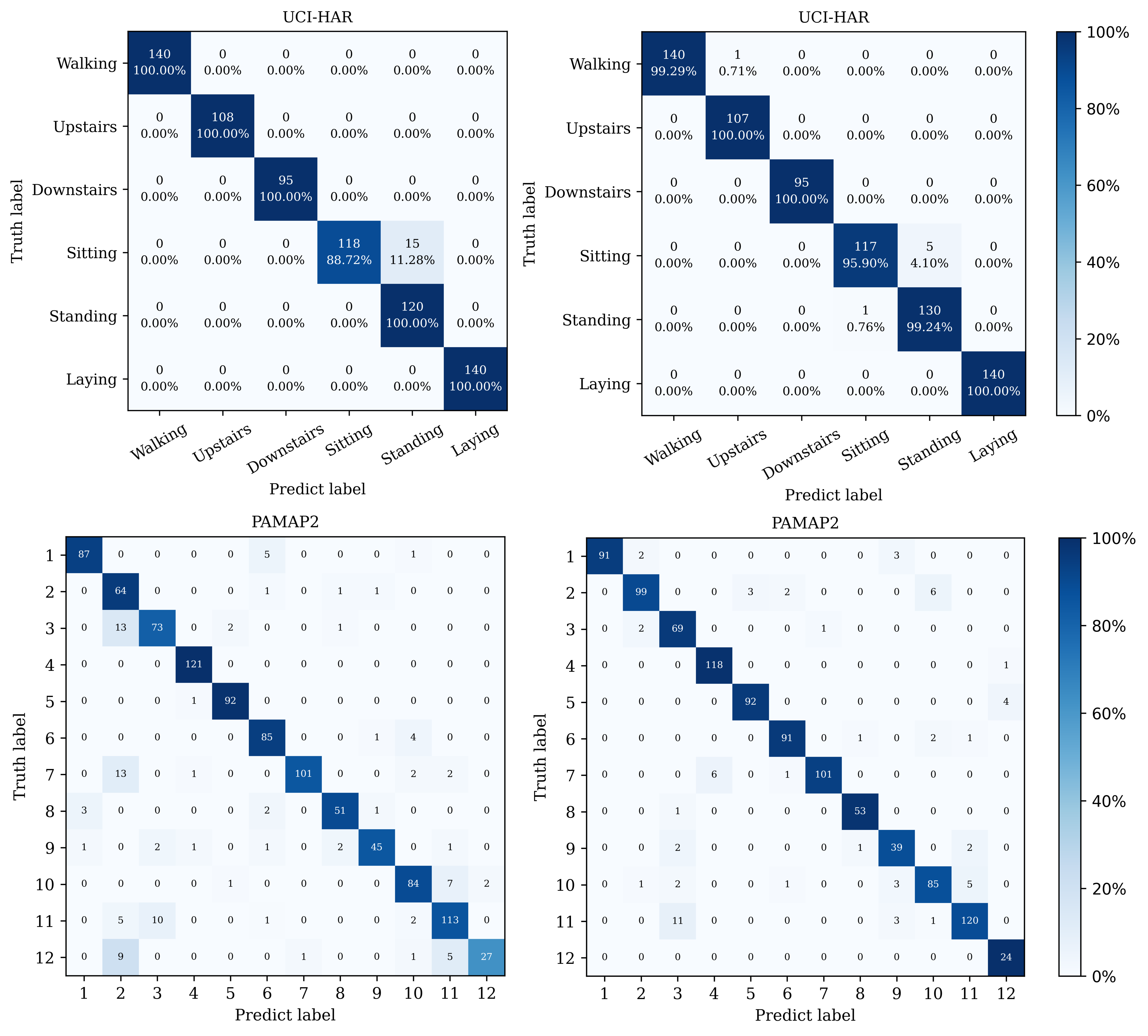}
    \caption{MSAP (left) and MAP-DM (right) confusion matrices on the UCI-HAR and PAMAP2 datasets}
    \label{fig4:hunxiao}
  \end{minipage}
  \hfill 
  \begin{minipage}[b]{0.53\textwidth}
    \includegraphics[width=\textwidth]{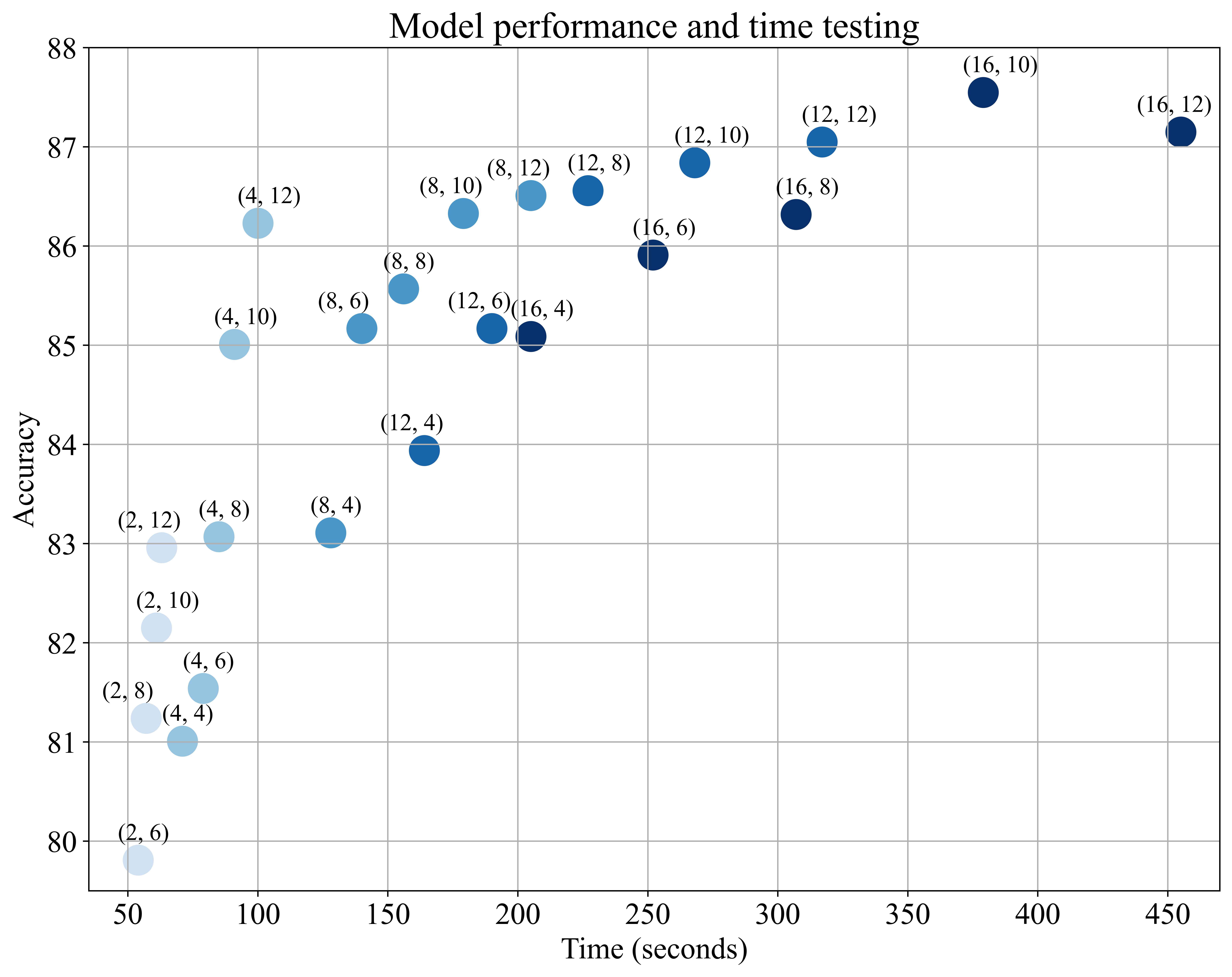}
    \caption{Performance of the MSAP-DM model on OPPORTUNITY datasets at different sizes}
    \label{fig:Time and Acc}
  \end{minipage}
\end{figure}
Figure \ref{fig4:hunxiao}. shows the confusion matrix of MSAP (left) and MSAP-DM (right) on UCI-HAR and PAMAP2 datasets. MSAP-DM achieves higher accuracy on more categories and effectively reduces the problems caused by confusing categories. For example, MSAP has high confusion between the "sitting" and "standing" categories on UCI-HAR, resulting in an accuracy of 88.72\% for "sitting", while MSAP-DM optimizes this to some extent. The accuracy of the "Sitting" category also reached 95.90\%, and the same situation also appeared between the "2", "3", "10" and "11" categories on PAMAP2. Compared with MSAP, the accuracy rate of the "MSAP-DM" was significantly improved due to its stronger ability to distinguish complex tasks.

\subsubsection{Model performance and time test}
It is worth noting that MSAP-DM achieves very excellent results on the OPPORTUNITY dataset, but these performance improvement is worth its resource consumption? For the MSAR-DM model, the most important factors affecting model performance and efficiency are Scales and Width, and there is a certain correlation trend between them and model performance. In order to explore this problem, we set up a network scale experiment on OPPORTUNITY to study the above relationship, and optimized the parameters (Scales and width) that mainly affect the model size in MSAP-DM.Figure \ref{fig:Time and Acc} articulates the relationship between MSAP-DM network size and performance and time consumption, as tested on the OPPORTUNITY dataset. It is evident that both the quantity of Scales and the Width of the Scales substantially influence the model's performance. For instance, setting scales to 12 or 16 yields high accuracy - a balancing act is needed. On the other hand, configurations such as (4, 12) offer comparable performance to larger models (less than 0.5\% difference), without significantly amplifying the cost. As a result, we prioritize parameter combinations that offer significant practical value. 
 
\subsection{Comparison with Existing Work}
We have conducted a comparative analysis between the MSAP-DM models and several corresponding models in the HAR domain. These models were re-implemented in accordance with our standard methodology. The results derived from our experiments are captured in Table \ref{tab7:Comparative Test}. It is salient to note that, when juxtaposed with other models, MSAP-DM delivered superior results.

\begin{table}
\centering
\caption{Comparison of related work}
\label{tab7:Comparative Test}
\begin{tabular}{ccccc} 
\toprule
{\textbf{Model}}&\textbf{ PAMAP2} & \textbf{WISDM }& \textbf{OPPORT}.& \textbf{UCI-HAR}\\
\cmidrule(r){1-5}
CNN \citet{M.Zeng_and_Nguyen}& 90.96
& 93.31 & 82.15 & 92.39 \\

LSTM \citet{L.MinhDang}& 91.51
& 96.71 & 81.65 & 95.52 \\

LSTM-CNN \citet{K.XiaJ.Huang}& 91.43
& 95.90 & 77.64 & 97.01 \\

CNN-GRU \citet{N.Dua_S.N.Singh}& 91.58
& 94.95 & 79.85 & 95.11 \\

SE-Res2Net \citet{S.-H.Gao_res2net}& 92.25
& 95.52 & 82.15 & 96.60 \\
 ResNeXt \citet{mekruksavanich2022deep}& 90.52& 96.67& 79.15& 95.38\\

Gated-Res2Net \citet{C.Yang}& 91.94& 97.02 & 81.51 & 96.31 \\

Rev-Att \citet{R.Pramanik}& 91.87& 97.46& 83.77& 95.53\\
 
HAR-CE\citet{CE}& 93.05& 97.72& 85.26& 98.89
\\
 
ELK \citet{ELK}& 92.76& 98.05& 82.80& 99.64
\\

DanHAR \citet{DanHAR}& 92.00& 95.57& 83.64& 96.21\\

\textbf{MSAP-DM}& \textbf{93.33}& \textbf{98.24}& \textbf{86.23}& \textbf{99.05}\\
\bottomrule
\end{tabular}\end{table}

\begin{figure}[htbp]
  \centering
  \begin{minipage}[b]{0.51\textwidth}
    \includegraphics[width=\textwidth]{pic/fenxi.pdf}
    \caption{Model performance analysis (Accuracy, F1-macro, F1-weighted)}
    \label{fig:Analysis}
  \end{minipage}
  \hfill 
  \begin{minipage}[b]{0.48\textwidth}
    \includegraphics[width=\textwidth]{pic/MSAPdelay.pdf}
    \caption{Inference delay of each model}
    \label{fig:inference delay}
  \end{minipage}
\end{figure}

Notably, MSAP-DM exhibits a significant improvement in accuracy compared to classical models.Furthermore, when comparing with state-of-the-art models, MSAP still maintains its leading position. Figure \ref{fig:Analysis} visually compares the performance and efficiency between MSAP-DM and advanced models. It is evident that MSAP-DM outperforms both multi-scale models Res2Net and SE-Res2Net by nearly 2\% and 1.5\% respectively. Additionally, compared to more advanced grouping topology network ResNeXt and traditional residual model, MSAP-DM demonstrates superior performance without increasing model complexity. In terms of efficiency, the classical model exhibits a remarkably high level of performance, enabling swift completion of computation and data processing tasks. However, its limited capability to handle complex tasks or large datasets diminishes its value as a reference point. Conversely, in our advanced model, we have optimized the DRSN interlayer structure which empowers MSAP-DM to outperform other models without significant time costs. 

\begin{table*}
\centering
\caption{Complexity analysis. }
\label{tab:Complexity analysis}
\begin{tabular}{ccccccc}
\toprule
\multirow{2}{*}{\textbf{Model}}& \multicolumn{3}{c}{WISDM} & \multicolumn{3}{c}{UCI} \\
\cmidrule(r){2-4}\cmidrule(r){5-7}
  & \textbf{\textbf{Time/ms}} & \textbf{\textbf{Memory}} & \textbf{\textbf{Param.}} & \textbf{\textbf{Time/ms}} & \textbf{\textbf{Memory}} & \textbf{\textbf{Param.}} \\
\cmidrule(r){1-1}\cmidrule(r){2-4}\cmidrule(r){5-7}
CNN & 23.78 & 766.23 & 1.04E+05 & 31.276 & 754.41 & 1.11E+05 \\
LSTM & 152.03 & 759.80 & 1.65E+05 & 235.30 & 746.08 & 2.75E+05 \\
CNN-GRU & 93.17 & 787.05 & 4.38E+06 & 124.89 & 777.69 & 4.38E+06 \\
LSTM-CNN & 545.74 & 823.00 & 2.12E+06 & 415.96 & 773.64 & 2.12E+06 \\
SE-Res2. & 443.32 & 787.67 & 1.60E+06 & 577.68 & 770.69 & 1.60E+06 \\
ResNeXt & 4520.88 & 889.78 & 2.20E+07 & 5982.67 & 856.02 & 2.20E+07 \\
Rev-Att & 458.49 & 862.03 & 4.33E+05 & 229.42 & 745.16 & 2.75E+05 \\
Gated-Res2. & 449.57 & 773.84 & 1.60E+06 & 577.08 & 762.38 & 1.60E+06 \\
CNN\_CE & 142.12 & 788.73 & 4.17E+05 & 405.15 & 829.11 & 4.31E+05 \\
ELK & 191.63 & 782.70 & 2.41E+05 & 831.19 & 753.88 & 2.50E+05 \\
Dan & 465.88 & 790.66 & 2.35E+06 & 2243.91 & 785.06 & 2.55E+06 \\
MSAP & 507.20 & 854.55 & 8.86E+06 & 667.99 & 881.30 & 8.86E+06 \\
\bottomrule
\toprule
\multirow{2}{*}{\textbf{Model}}& \multicolumn{3}{c}{PAMAP2} & \multicolumn{3}{c}{OPPO} \\
\cmidrule(r){2-4}\cmidrule(r){5-7}
  & \textbf{\textbf{Time/ms}} & \textbf{\textbf{Memory}} & \textbf{\textbf{Param.}} & \textbf{\textbf{Time/ms}} & \textbf{\textbf{Memory}} & \textbf{\textbf{Param.}} \\
\cmidrule(r){1-1}\cmidrule(r){2-4}\cmidrule(r){5-7}
CNN & 44.863 & 762.48 & 1.59E+05 & 14.563 & 744.52 & 1.21E+05 \\
LSTM & 325.87 & 752.19 & 3.04E+05 & 62.924 & 748.44 & 3.85E+05 \\
CNN-GRU & 159.56 & 768.95 & 4.38E+06 & 50.362 & 773.17 & 4.39E+06 \\
LSTM-CNN & 596.81 & 765.00 & 2.13E+06 & 84.83 & 788.69 & 2.13E+06 \\
SE-Res2. & 711.43 & 780.34 & 1.60E+06 & 165.90 & 772.41 & 1.61E+06 \\
ResNeXt & 8205.77 & 854.20 & 2.21E+07 & 1522.20 & 849.25 & 2.21E+07 \\
Rev-Att & 3524.24 & 1040.13 & 3.77E+06 & 204.91 & 801.97 & 3.80E+05 \\
Gated-Res2. & 814.93 & 784.70 & 1.60E+06 & 163.90 & 769.53 & 1.61E+06 \\
HAR\_CE & 3276.34 & 779.84 & 1.65E+06 & 1754.43 & 816.86 & 3.49E+06 \\
ELK & 5264,80 & 751.84 & 3.47E+05 & 2687.07 & 770.75 & 7.57E+05 \\
DanHAR & 14841.37 & 814.09 & 5.12E+06 & 7773.65 & 803.44 & 4.46E+06 \\
MSAP-DM & 1027.66 & 846.17 & 1.99E+07 & 785.91 & 856.05 & 7.96E+07 \\
\bottomrule

\end{tabular}

\end{table*}

\subsection{Actual deployment}

To demonstrate the feasibility and real-time detection performance of the model on embedded devices, we selected the Raspberry Pi 5 as the actual deployment platform. It is based on the Broadcom BCM2712 2.4GHz quad-core 64-bit Arm Cortex-A76 CPU with 8 GB RAM. The operating system used is Raspberry Pi OS, with Python version 3.11.2 and PyTorch version 2.2.2 as the deep learning framework. All models were trained on the WISDM, OPPORT, PAMAP2, and UCI-HAR datasets and imported into the embedded simulation platform in the same format for testing. The model inference tests consisted of two parts to simulate real-world conditions.

The first part of the test involved continuously inputting 100 pieces of data to simulate a scenario where a large amount of data is input simultaneously. The evaluation metrics included running time, average memory usage, and model parameters, as detailed in Table \ref{tab:Complexity analysis}. From the tables, it is evident that although our model has higher parameter counts and memory usage, the inference time has not increased significantly.The second part of the test involved inputting a single piece of data at a time to simulate continuous real-time monitoring. This test was repeated 100 times to reflect the model's stability. The evaluation metric was the single inference time for the entire test. The input data was uniformly segmented using a sliding window of 95\%. This means that as long as the model's single inference time is less than 5\% of the data time, we consider the model deployable on the embedded platform. This standard is shown as a red dashed line in figure \ref{fig:inference delay} .On the WISDM dataset, where the single data time is relatively long, all models completed inference within the required time. However, on the UCI dataset, which has shorter time intervals, the ResNeXt model struggled to meet the requirements. On datasets with even shorter time intervals and larger single data sizes, such as PAMAP2 and OPPORTUNITY, not only did ResNeXt fail to meet the requirements, but the DanHAR model also experienced increased inference time due to overheating of the embedded system. In contrast, our model met the time requirements on all datasets.

\section{Conclusion} \label{Conclusion}
In this study, we first explore the feature extraction capabilities of multi-scale convolutional neural networks and propose a multi-scale CNNs with an integrated attention feature purification module. Furthermore, we introduce a deep residual shrinkage network in HAR to reduce redundant features while adjusting its structure to be more compatible with HAR data. Experiments show that our method only adds a small amount of resource cost, but achieves huge performance improvements, surpassing all previous methods. It is worth mentioning that in this paper we propose an embedded test system suitable for the performance of current wearable devices and extensively verify the rationality of the existing HAR model, which enables the HAR method to not only run on high-performance devices , and can conduct simulations that are closer to reality. Our research makes a great contribution to the application of sensor-based HAR.
 
\bibliographystyle{plainnat}
\bibliography{neurips_2024}
\section{Appendix / supplemental material}
\subsection{Code}
All relevant code is in this anonymous github repository: https://anonymous.4open.science/r/MSAP-C345
\subsection{Limitations}
The limitations of this work are mainly reflected in the proposed embedded system. We try our best to simulate the real detection situation, but the actual situation of wearable devices will be more complex, and performance, power consumption and interference issues from other processes need to be considered. We plan to deploy this system on real civilian wearable devices in future work to solve this problem.

\end{document}